\documentclass{article}

\usepackage{microtype}
\usepackage{graphicx}
\usepackage{subfigure}
\usepackage{booktabs} 

\usepackage[utf8]{inputenc} 
\usepackage[T1]{fontenc}    
\usepackage{hyperref}       
\usepackage{url}            
\usepackage{booktabs}       
\usepackage{amsfonts}       
\usepackage{nicefrac}       
\usepackage{microtype}      

\usepackage{dsfont}    
\usepackage{todonotes}
\usepackage{amsmath,amssymb}
\usepackage{mathtools}
\usepackage{amsthm}
\usepackage{multirow}
\usepackage{listings}

\newcommand{\E}{\mathbb{E}}
\newcommand{\mean}[2]{\E_{#2} \! \left[ #1 \right]}
\newcommand{\vt}[1]{\boldsymbol{#1}}

\usepackage{hyperref}

\usepackage[accepted]{icml2021}

\icmltitlerunning{Automatic structured variational inference with cascading flows}

\begin{document}

\twocolumn[
\icmltitle{Automatic variational inference with cascading flows}



\icmlsetsymbol{equal}{*}

\begin{icmlauthorlist}
\icmlauthor{Luca Ambrogioni}{ru}
\icmlauthor{Gianluigi Silvestri}{op,ru}
\icmlauthor{Marcel van Gerven}{ru}
\end{icmlauthorlist}

\icmlaffiliation{ru}{Donders Centre for Cognition, Radboud University, Netherlands}
\icmlaffiliation{op}{One Planet, Netherlands}

\icmlcorrespondingauthor{Luca Ambrogioni}{l.ambrogioni@donders.ru.nl}

\icmlkeywords{Machine Learning, ICML}

\vskip 0.3in
]



\printAffiliationsAndNotice{\icmlEqualContribution} 

\begin{abstract}
The automation of probabilistic reasoning is one of the primary aims of machine learning. Recently, the confluence of variational inference and deep learning has led to powerful and flexible automatic inference methods that can be trained by stochastic gradient descent.  In particular, normalizing flows are highly parameterized deep models that can fit arbitrarily complex posterior densities. However, normalizing flows struggle in highly structured probabilistic programs as they need to relearn the forward-pass of the program. Automatic structured variational inference (ASVI) remedies this problem by constructing variational programs that embed the forward-pass. Here, we combine the flexibility of normalizing flows and the prior-embedding property of ASVI in a new family of variational programs, which we named cascading flows. A cascading flows program interposes a newly designed highway flow architecture in between the conditional distributions of the prior program such as to steer it toward the observed data. These programs can be constructed automatically from an input probabilistic program and can also be amortized automatically. We evaluate the performance of the new variational programs in a series of structured inference problems. We find that cascading flows have much higher performance than both normalizing flows and ASVI in a large set of structured inference problems. 
\end{abstract}

\section{Introduction}
The aim of probabilistic programming is to provide a fully automated software system for statistical inference on arbitrary user-specified probabilistic models (also referred to as probabilistic programs)~\cite{milch20071, sato1997prism, kersting20071, pfeffer2001ibal, park2005probabilistic, goodman2012church, wingate2011lightweight, patil2010pymc, dillon2017tensorflow, bingham2019pyro, tran2017deep, tran2016edward}. When the probabilistic programs are built from differentiable components, stochastic variational inference (VI) by automatic differentiation offers an effective and computationally efficient solution~\cite{bishop2002vibes, kucukelbir2017automatic, tran2017deep}. However, the performance of stochastic VI depends strongly on the choice of a variational program (also known as variational family or variational guide)~\cite{bishop2003structured}. 

Until recently, automatic variational program construction was limited to simple mean field and multivariate normal coupling approaches~\cite{kucukelbir2017automatic}. These programs are highly constrained and exploit only minimally the structure of the original probabilistic program. For example, the multivariate normal approach only depends on the number of variables and their support. More recently, normalizing flows revolutionized the field of variational inference by offering a highly flexible parametric model for complex multivariate distributions. Normalizing flows can be easily used to construct automatic variational programs by mapping all the variables to a spherical normal latent space through a learnable diffeomorphism~\cite{rezende2015variational, papamakarios2019normalizing}. While highly flexible, the normalizing flow approach still uses only minimal information concerning the input probabilistic program. This leads to poor performance in highly structured probabilistic programs with complex computational flows~\cite{ambrogioni2020automatic}. For example, consider an inference problem defined by a highly parameterized natural language model ~\cite{brown2020language}. A typical problem of this kind is controlled language generation, where the goal is to steer the model to general text conditioned on its past output and future targets (e.g. the presence of some specific words). The controlled network can be formulated in probabilistic programming terms as a posterior distribution that is parameterized by the variational program. In a problem like this, most of the complexity of the posterior is encoded in the structure of the prior probabilistic program (the language model). The model just needs to be steered in the right direction. However, normalizing flow completely disregards this structure and has to relearn all the structure embedded in the prior. 

This problem was discussed in a recent work that introduced a form of automatic structured variational inference (ASVI) with an automated variational program that incorporates the prior probabilistic program as a special case. ASVI has been shown to have very good performance in time series analysis and deep learning problems in spite of being very parsimoniously parameterized compared to normalizing flows~\cite{ambrogioni2020automatic}. However, the convex-update families used in ASVI have two main limitations: I) they cannot model statistical dependencies arising from colliders in the graphical model (the "explaining away" effect) and II) they constrain the conditional posterior distributions to be in the same family as the prior. In this paper we integrate ASVI and normalizing flows to obtain a new form of automatic VI, which we refer to as {\em cascading flows} (CF). The approach incorporates the forward-pass of the probabilistic program while being capable of modeling collider dependencies and arbitrarily complex conditional distributions. To this end, we make use of a novel highway flow architecture. CF respects the following design principles:
\begin{itemize}
    \item Automation: It can be constructed using a fully automatic algorithm that takes the prior probabilistic program as input. 
    \item Locality: It is constructed by locally transforming each conditional distribution in the conditional prior so that it inherits the graphical structure of the input program (with potentially some extra coupling). This ensures scalability and modularity.
    \item Prior information flow: It embeds the forward-pass of the probabilistic program.
\end{itemize}





\section{Related work}
Automatic algorithms for VI date back to VIBES~\cite{bishop2002vibes, bishop2003structured}. These older works are based on self-consistency equations and variational message passing~\cite{winn2005variational}. The use of gradient-based automatic VI for probabilistic programming was pioneered in~\cite{wingate2013automated} using a mean field approach. Automatic differentiation VI (ADVI) perfected this approach by exploiting the automatic-differentiation tools developed in the deep learning community~\cite{kucukelbir2017automatic}. ADVI uses fixed invertible transformations to map each variable into an unbounded domain where the density can be modeled with either univariate of multivariate normal distributions. Structured stochastic VI~\cite{hoffman2015structured} exploits the local conjugate update rules of locally conjugate models. This has the major benefit of preserving the local structure of the probabilistic program but it severely limits the class of possible input programs since local conjugacy is a stringent condition. Automatic structured VI (ASVI) lifts this constraint as it applies a trainable convex-update rule to non-conjugate links~\cite{ambrogioni2020automatic}. Several other papers introduced new techniques for constructing structured variational programs. Copula VI introduces the use of vine copulas to model the statistical coupling between the latent variables in the program~\cite{tran2015copula}. Similarly, hierarchical VI uses auxiliary latent variables to achieve the same goal~\cite{ranganath2016hierarchical}. Neither of those approaches are truly automatic however, as they do not prescribe a unique way to construct the coupling functions. 

Normalizing flows are very expressive and highly parameterized models that use deep learning components to approximate complex multivariate densities~\cite{rezende2015variational}. Normalizing flows can approximate arbitrarily complex densities and can therefore be turned into powerful automatic VI methods~\cite{rezende2015variational, kingma2016improved}. However, conventional normalizing flows do not exploit the input probabilistic program. Structured conditional continuous normalizing flows are a newly introduced class of flows constrained to have the same conditional independence structure of the true posterior~\cite{weilbach2020structured}. However, these flow architectures only inherit the graphical structure of the input program, without incorporating the forward pass (i.e. the form of the conditional distributions and link functions). 

Modern deep probabilistic programming frameworks provide automatic implementation of many of variational programs. For example, PyMC~\cite{patil2010pymc} provides an implementation of ADVI, Pyro~\cite{bingham2019pyro} implements the automatic construction of mean field, multivariate normal and normalizing flow programs (i.e. variational guides) and TensorFlow probability also offers an automatic implementation of ASVI~\cite{dillon2017tensorflow, ambrogioni2020automatic}. 

Our new highway flow architecture is a key component of our new approach and it is inspired by highway networks~\cite{srivastava2015highway}. To the best of our knowledge, highway flows are the first form of residual-like flows with a tractable Jacobian determinant that can be expressed in closed form. Existing residual flows express the log determinant of the Jacobian as an infinite series that needs to be estimated using a “Russian roulette” estimator~\cite{chen2019residual}. We achieve tractability by applying a highway operator to each single layer of the network.

Our use of auxiliary variables is similar to the approach adopted in~\cite{dupont2019augmented} and~\cite{weilbach2020structured}. The bound we use for training the augmented model was initially derived in~\cite{ranganath2016hierarchical}. To the best of our knowledge we are the first to exploit the statistics of the auxiliary variables to implement inference amortization. 

Structured variational programs have the most clear benefits in timeseries analysis problems~\cite{eddy1996hidden, foti2014stochastic, johnson2014stochastic, karl2016deep, fortunato2017bayesian}. Our approach, together with ASVI, differs from the conventional timeseries approach by exploiting the structure of both temporal and non-temporal variables in the model.


\section{Preliminaries}
In this section we introduce the notation used in the rest of the paper and discuss in detail the methods that form the basis for our work. 

\subsection{Differentiable probabilistic programs}
We denote the values of (potentially multivariate) random variables using lowercase notation and arrays of random variables using boldface notation. We consider probabilistic programs constructed as a chain of conditional probabilities over an array of random variables $\vt{x} = (x_1, \ldots, x_N)$. We  denote the array of values of the parents of the $j$-th variable as $\vt{\pi}_j \subseteq \{x_i\}_{i \neq j}$, which is a sub-array of parent variables such that the resulting graphical model is a directed acyclic graph. Using this notation, we can express the joint density of a probabilistic program as follows:
\begin{align}
    p \left(\vt{x} \right) &= \prod_{j=1}^N  \rho_j \left(x_j  \mid \theta_j(\vt{\pi}_j) \right)~,
    \label{eq: probabilistic program}
\end{align} 
where $\theta_j(\vt{\pi}_j)$ is a link function that maps the values of the parents of the $j$-th variable to the parameters of its density $\rho_j \left(\cdot \mid \cdot \right)$. For example, if the density is Gaussian the link function outputs the value of its mean and scale given the values of the parents. Note that $\theta_j$ is a constant when the array of parents is empty.

\subsection{Convex-update variational programs}
An automatic structured variational inference method provides an algorithm that takes a probabilistic program as input and outputs a variational program $q(\vt{x})$. Convex-update variational programs~\cite{ambrogioni2020automatic} are defined by the following transformation:
\begin{align}
   p(\vt{x}) &\underset{\text{CU}}{\mapsto}  q(x) = \prod_j^N \mathcal{U}_{{\lambda}_j}^{{\alpha}_j} \rho_j \left(x_j  \mid \theta_j(\vt{\pi}_j) \right)~,
    \label{eq: ASVI}
\end{align} 
with convex-update operator 
\begin{align}
    \mathcal{U}_{{\lambda}_j}^{{\alpha}_j} \rho_j \left(x_j  \mid \theta_j(\vt{\pi}_j) \right) = \rho_j (x_j \mid & {\lambda}_j \odot \theta_j(\vt{\pi}_j) \\ \nonumber
    & + (1 - {\lambda}_j) \odot {\alpha}_j )~,
\end{align}
where ${\lambda}_j$ and ${\alpha}_j$ are (potentially vector-valued) learnable parameters and $\odot$ is the element-wise product. This reduces to the prior probabilistic program for ${\lambda}_j \rightarrow 0$ and to a mean field variational program for ${\lambda}_j \rightarrow 1$. Convex-update ASVI has the advantage of preserving the forward-pass of the probabilistic program, which is often a major source of statistical coupling in the posterior distribution. The variation program can be trained to fit the posterior by minimizing the evidence lower bound (ELBO) by stochastic gradient descent (SGD) on the differentiable parameters $\lambda_j$ and ${\alpha}_j$. 

\subsection{Variational inference with normalizing flows}
Normalizing flows express the density of arbitrary multivariate distributions as the push forward of a known base distribution through a differentiable invertible function (i.e. a diffeomorphism). The base distribution is usually a spherical normal distribution, hence the "normalizing" in the name. Consider a $d$-variate probability density $p_0(z)$ and a diffeomorphism $\Psi: \mathds{R}^d \rightarrow \mathds{R}^d$. We can derive the density of the new random variable $x = \Psi(z)$ using the change of variable formula
\begin{equation}\label{eq: change of variable}
    p_X(x) = |\det{J}(\Psi^{-1}(x))| p_0(\Psi^{-1}(x))~,
\end{equation}
where $\Psi^{-1}$ is the inverse transformation and $J(z)$ is the Jacobi matrix of $\Psi$. We can write this more compactly using the push-forward operator $\mathcal{T}_\Psi$ associated to $\Psi$ that maps the density $p_0(z)$ to the density $p_X(x)$ given in Eq.~\ref{eq: change of variable}: $p_X(x) = \mathcal{T}_\Psi \left[ p_0(\cdot) \right](x)$. 

Now consider a flexible family of functions $\Psi^w(x)$ parameterized by the "weights" $w$. We can use Eq.~\ref{eq: activation} to approximate arbitrarily complicated target densities by training $w$ by SGD. Given a probabilistic program $p(x,y)$ with $y$ observed, we can approximate the posterior $p(x\mid y)$ by minimizing the evidence lower bound:
\begin{equation}
    \text{ELBO}(w) = \mean{\log{\frac{p(x, y)}{|\det{J}_w(\Psi^{-1}(x))| p_0(\Psi^{-1}(x))}}}{w}
\end{equation}
where the expectation is taken with respect to the transformed distribution $\mathcal{T}_{\Psi^w} \left[p_0(\cdot) \right](x)$. Since $\Psi$ is differentiable, we can obtain the reparameterization gradient for the ELBO as follows:
\begin{equation}\label{eq: rep ELBO gradient}
    \nabla_w \text{ELBO}(w) = \mean{ \nabla_w \log{\frac{p(\Psi^w(z), y)}{|\det{J}_w(z)| p_0(z)}}}{0}
\end{equation}
where the expectation is now taken with respect to the fixed distribution $p_0(z)$. In the following we refer to this approach as \emph{global flow} (GF) since the transformation is applied synchronously to all variables in the model. 

\section{Variational inference with cascading flows}
Now that we have covered the relevant background, we will introduce our new approach that combines ASVI with normalizing flows. 

\subsection{Cascading flows variational programs}
The convex-update parameterization poses strong constraints on the form of the conditional posteriors. On the other hand, normalizing flow programs do not embed the potentially very complex structure of the input program. In this paper we combine the two approaches by replacing the convex-update operator in Eq.~\ref{eq: ASVI} with the push-forward operator used in normalizing flows. More precisely, consider a family of diffeomorphisms $\vt{\Psi}^{\vt{w}} = (\Psi_1^{w_1},\dots,\Psi_N^{w_N})$ depending on the array of parameters $\vt{w} = (w_1,\dots,w_N)$. Using these transformations, we can define the \emph{cascading flows} variational program associated with a program defined by Eq.~\ref{eq: probabilistic program}:
\begin{align}
   p(\vt{x}) \underset{\text{CF}}{\mapsto}  q_{\vt{w}} \left(\vt{x} \right) &= \prod_j^N \mathcal{T}_j^{\vt{w}} \left[ \rho_j \left(\cdot  \mid \theta_j(\vt{\pi}_j) \right) \right] (x_j)~,
\end{align}
where we denoted the push-forward operator induced by the diffeomorphism $\Psi_j^{w_j}$ (as defined by Eq.~\ref{eq: change of variable}) as $\mathcal{T}_j^{\vt{w}}$ to simplify the notation. A cascading flows variational program can be trained by SGD using gradient estimators of the ELBO with the reparameterization given in Eq.~\ref{eq: rep ELBO gradient}. 

\begin{figure}[ht]
    \centering
    \includegraphics[width=0.45\textwidth]{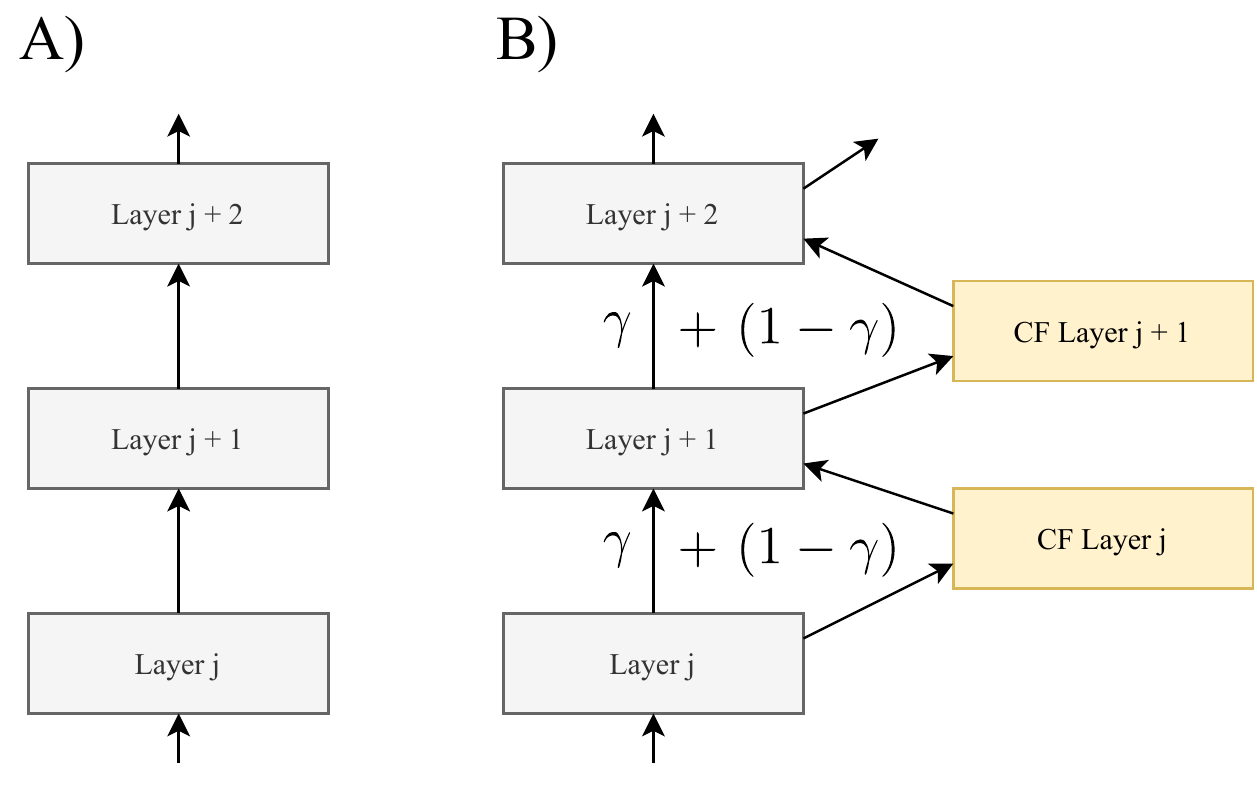}
   \caption{Diagram of a cascading flows architecture. A) Detail of architecture of an input probabilistic program. B) Detail of the associated cascading flows architecture. }
    \label{fig: eq: cascading flow}
\end{figure}

The name cascading flows comes from the fact that, since the transformation is performed locally for each conditional distribution, its effects cascade down the variational model through the (transformed) conditional dependencies of the prior program. However, in order to preserve the forward-pass of the probabilistic program it is crucial to use a parameterized diffeomorphism that can be initialized around the identity function and whose deviation from the identity can be easily controlled. Specifically, we opt for transformations of the following form:
\begin{equation}\label{eq: highway transformation}
    \Psi_j^{w_j}(x) = \gamma x + (1 - \gamma) f_j(x; w_j) \,~,
\end{equation}
where $\gamma \in (0,1)$. The resulting architecture of a cascading flows variational program is visualized in Fig~\ref{fig: eq: cascading flow}. From the diagram it is clear that a cascading flow is constructed by inserting transformation layers within the architecture of the probabilistic program. Transformations of the form given in Eq.~\ref{eq: highway transformation} with tractable Jacobians are not currently present in the normalizing flow literature. Therefore in the next subsection we will introduce a new flow architecture of this form, which we named \emph{highway flow}.

\subsection{Highway flow architecture}

\begin{figure}[ht]
    \centering
    \includegraphics[width=0.4\textwidth]{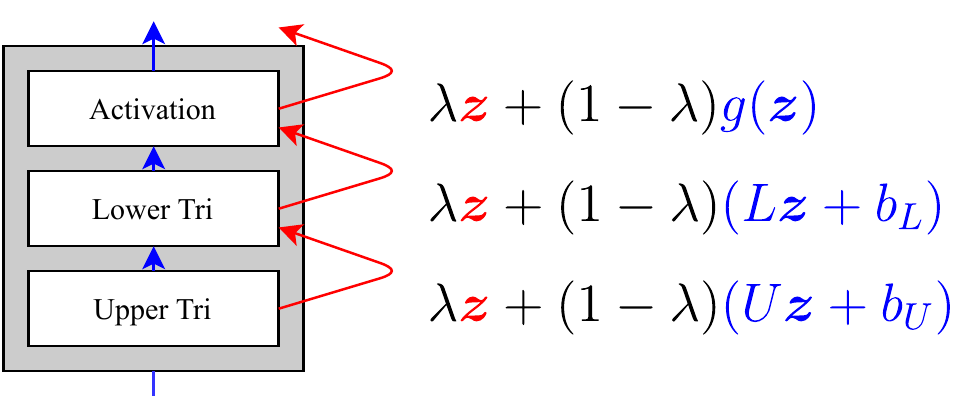}
   \caption{Diagram of a highway flow block.}
    \label{fig: highway}
\end{figure}
Here we introduce highway flow blocks gated by a learnable scalar gate parameter. As also visualized in Fig.~\ref{fig: highway}, a highway flow block is comprised by three separate highway layers:
\begin{enumerate}
    \item Upper triangular highway layer:
\begin{align}
    &l_U({z}; U, \lambda) = \lambda {z} + (1 - \lambda) \left(U {z} + b_U\right) \label{eq: L_u} \\
    &\log{\det{J_U}} = \sum_k \log{\left(\lambda + (1 - \lambda) U_{kk} \right)}
\end{align}

\item Lower triangular layer:
\begin{align}
    &l_L(\vt{z}; L, \lambda) = \lambda {z} + (1 - {\lambda}) \left(L {z} + b_L\right) \label{eq: L_l}\\
    & \log{\det{J_L}} = \sum_k \log{\left(\lambda + (1 - \lambda) L_{kk} \right)}
\end{align}

\item Highway activation functions:
\begin{align}
    &f(\vt{z}; {\lambda}) = {\lambda} {z} + (1 - {\lambda}) g({z}) \label{eq: activation}\\
    & \log{\det{\frac{\text{d} f(x_k)}{\text{d} x}}} = \sum_k \log{\left(\lambda + (1 - \lambda) \frac{\text{d} g(x_k)}{\text{d} x} \right)}
\end{align}
\end{enumerate}
where $U$ is an upper-triangular matrix with positive-valued diagonal, $L$ is a lower-triangular matrix with ones in the diagonal and $g(x)$ is a differentiable non-decreasing activation function. A highway flow layer is a composition of these three types of layers:
\begin{equation}
    l_h (\cdot) =  f \circ l_L \circ  l_U (\cdot) \,.
\end{equation}

A highway network is a sequence of $M$ highway layers with a common gate parameter and different weights and biases. Note that a highway flow can be expressed in the form given in Eq.~\ref{eq: highway transformation} by defining $\gamma$ as follows: 
\begin{equation}
    \gamma = \lambda^{3 M}~,
\end{equation}
which is clearly equal to one (zero) when $\lambda$ is equal to one (zero).

\subsection{Auxiliary variables and infinite mixtures of flows}
The expressiveness of a normalizing flow is limited by the invertible nature of the transformation. This is particularly problematic for low-dimensional variables since the expressivity of many flow architectures depends on the dimensionality of the input~\cite{papamakarios2019normalizing}. This limitation can be a major shortcoming in cascading flows as they are particularly useful in highly structured problems characterized by a large number of coupled low-dimensional variables. Fortunately, we can remedy this limitation by introducing a set of auxiliary variables. For each variable $x_j$, we create a $D$-dimensional variable $\vt{\epsilon}_j$ following a base distribution $p_j(\vt{\epsilon}_j)$. We can now use an augmented diffeomorphism $\hat{\Psi}_j^{w_j}(x_j, \vt{\epsilon}_j)$ and define the joint posterior over both $x_j$ and $\vt{\epsilon}_j$:
\begin{align}
    q \left(x_j, \epsilon_j  \mid \vt{\pi}_j \right) &=  \hat{\mathcal{T}}_j^{\vt{w}} \left[ \rho_j \left(\cdot  \mid \theta_j(\vt{\pi}_j) \right) p_j(\cdot) \right](x_j, \epsilon_j) \,,
\end{align}
where the push-forward operator now transforms the (independent) joint density of $x_j$ and $\epsilon_j$. The conditional variational posterior is then obtained by marginalizing out $\vt{\epsilon}_j$:
\begin{align}
    q \left(x_j  \mid \vt{\pi}_j \right) &= \int q \left(x_j, \vt{\epsilon}_j  \mid \vt{\pi}_j \right) \text{d} \vt{\epsilon}_j \,.
\end{align}
This infinite mixture of flows is much more capable of modeling complex and potentially multi-modal distributions. 

The use of latent mixture models for VI was originally introduced in~\cite{ranganath2016hierarchical}. Given a set of observations $\vt{y}$, the ELBO of any mixture variational probabilistic program can be lower bounded using Jensen's inequality: 
\begin{align} \label{eq: mixture ELBO}
    &\mean{\log{\frac{p(\vt{x}, \vt{y})} {\int q \left(\vt{x}, \vt{\epsilon} \right) \text{d} \vt{\epsilon}}}} {\vt{x}}\geq \underbrace{ \mean{\log{\frac{p(\vt{x}, \vt{y}) r(\vt{\epsilon})}{q \left(\vt{x}, \vt{\epsilon} \right)}}} {\vt{x}, \vt{\epsilon}}}_\text{Augmented ELBO}~, 
\end{align}
where $q(\vt{\epsilon}) = \int q \left(\vt{x}, \vt{\epsilon} \right) \text{d} \vt{x}$ is the marginal variational posterior and $r(\vt{\epsilon})$ is an arbitrarily chosen distribution over the auxiliary variables. This result justifies the use of the augmented ELBO  for training the variational posterior. The gap between the mixture ELBO (Eq.~\ref{eq: mixture ELBO}) and the augmented ELBO can be reduced by parameterizing $r(\vt{\epsilon})$ and training its parameters by minimizing the augmented ELBO such as to match $q(\vt{\epsilon})$.

The highway flow architecture can be easily adapted to the augmented variable space. Since there is no need to gate the flow of the auxiliary variables, the scalar $\lambda$ in Eq.~\ref{eq: L_u}, Eq.~\ref{eq: L_l} and Eq.~\ref{eq: activation} should be replaced by the vector $\vt{l}$ whose entries corresponding to the original variables are equal to $\lambda$ while the entries corresponding to the auxiliary variables are equal to zero. 

\subsection{Modeling collider dependencies with auxiliary coupling}

\begin{figure}[ht]
    \centering
    \includegraphics[width=0.4\textwidth]{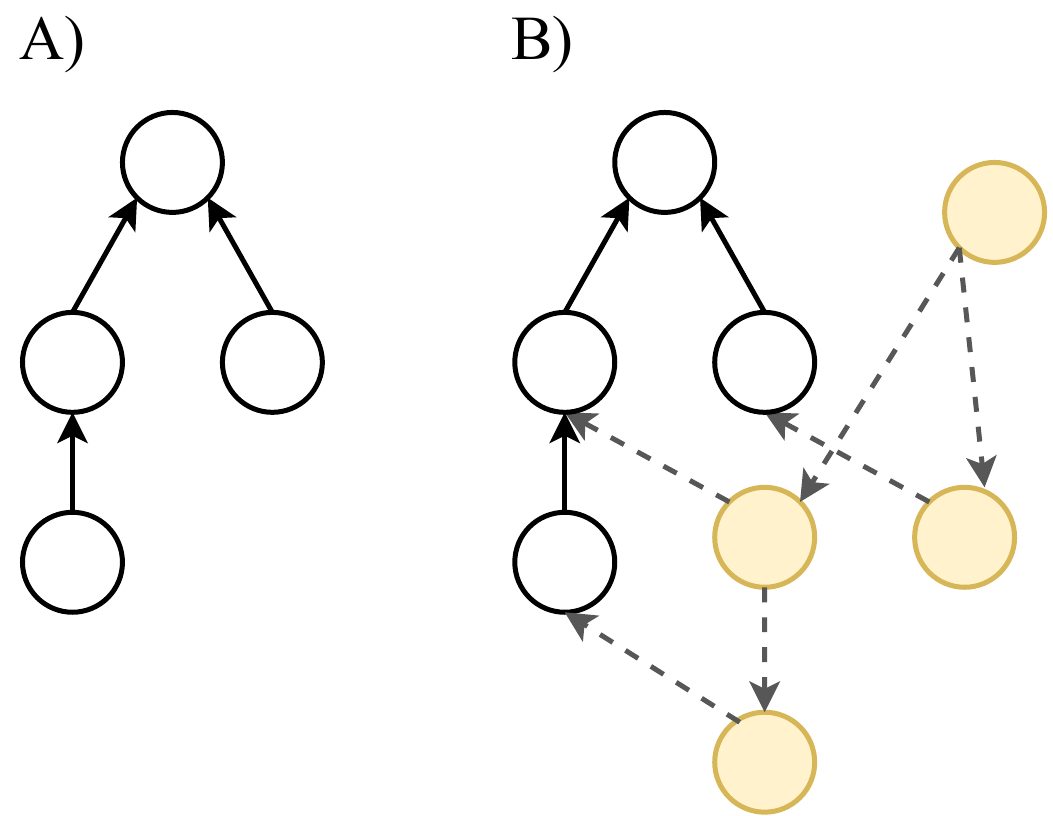}
   \caption{Backward auxiliary coupling. A) Example graphical model. B) Cascading flows model coupled with reversed local linear Gaussian model.}
    \label{fig: colliders}
\end{figure}

\begin{table*}[ht] \label{tab: timeseries}
\footnotesize
\centering
\caption{Predictive and latent log-likelihood (forward KL) of variational timeseries models. Error are SEM estimated over 10 repetitions.}
\begin{tabular}{llcccccc}
\centering
\small
{} & {} & {CF} & {ASVI} & {MF} & {GF} & {MVN} & {CF (non-res)} \\ \midrule
{BR-r}      & Pred & $-2.27 \pm 0.26$ & $\boldsymbol{-2.23 \pm 0.21}$ & $-3.79 \pm 0.82$ & $-2.81 \pm 0.56$ & $-2.88 \pm 0.53$ & $-3.33 \pm 0.65$ \\
& Latent & 
 $-1.48 \pm 0.19$ & $\boldsymbol{-1.45 \pm 0.14}$ & $-4.02 \pm 0.63$ & $-2.41 \pm 0.52$ & $-2.02 \pm 0.48$ & $-3.63 \pm 0.74$ \\ \midrule
{BR-c}     & Pred & $1.61 \pm 0.18$ &
     $1.45 \pm 0.14$ &
     $1.04 \pm 0.03$ &
     $\boldsymbol{2.00 \pm 0.29}$ &
     $1.02 \pm 0.03$ &
     $1.31 \pm 0.18$ \\
& Latent &
     $\boldsymbol{-1.53 \pm 0.21}$ &
     $-1.55 \pm 0.19$ &
     $-5.78 \pm 0.89$ &
     $-2.06 \pm 0.53$ &
     $-2.82 \pm 0.77$ &
     $-5.07 \pm 0.85$ \\ \midrule
{LZ-r}      & Pred & $\boldsymbol{-2.89 \pm 0.17}$ &
     $-4.48 \pm 0.60$ &
     $-8.26 \pm 0.28$ &
     $-8.03 \pm 0.37$ &
     $-8.24 \pm 0.29$ &
     $-8.25 \pm 0.27$ \\ 
& Latent &
     $\boldsymbol{-2.39 \pm 0.45}$ &
     $-4.38 \pm 0.67$ &
     $-10.28 \pm 0.18$ &
     $-9.44 \pm 0.20$ &
     $-9.45 \pm 0.22$ &
     $-10.00 \pm 0.18$ \\ \midrule
{LZ-c}     & Pred &
     $\boldsymbol{5.10 \pm 0.52}$ &
     $0.92 \pm 0.03$ &
     $0.90 \pm 0.003$ &
     $0.86 \pm 0.15$ &
     $0.89 \pm 0.001$ &
     $0.88 \pm 0.04$ \\
& Latent &
     $\boldsymbol{-4.19 \pm 0.66}$ &
     $-7.47 \pm 0.30$ &
     $-9.89 \pm 0.19$ &
     $-8.71 \pm 0.32$ &
     $-8.58 \pm 0.34$ &
     $-9.59 \pm 0.29$ \\ \midrule
{PD-r}      & Pred & $\boldsymbol{-3.19 \pm 0.22}$ &
     $-3.25 \pm 0.11$ &
     $-4.42 \pm 0.22$ &
     $-3.84 \pm 0.28$ &
     $-4.30 \pm 0.22$ &
     $-4.29 \pm 0.25$  \\
& Latent &
     $\boldsymbol{-2.32 \pm 0.19}$ &
     $-3.14 \pm 0.12$ &
     $-9.12 \pm 0.29$ &
     $-4.16 \pm 0.33$ &
     $-7.72 \pm 0.30$ &
     $-8.27 \pm 0.36$  \\ \midrule
{PD-c} & Pred & $\boldsymbol{1.97 \pm 0.07}$ &
     $1.65 \pm 0.06$ &
     $0.86 \pm 0.003$ &
     $01.07 \pm 0.02$ &
     $1.09 \pm 0.02$ &
     $0.96 \pm 0.01$ \\
& Latent &
     $\boldsymbol{-2.77 \pm 0.18}$ &
     $-3.09 \pm 0.15$ &
     $-8.40 \pm 0.43$ &
     $-6.20 \pm 0.40$ &
     $-7.45 \pm 0.42$ &
     $-8.41 \pm 0.43$ \\ \midrule
{RNN-r}      & Pred & $-1.68 \pm 0.05$ &
     $-2.30 \pm 0.18$ &
     $-5.20 \pm 0.94$ &
     $\boldsymbol{-1.60 \pm 0.09}$ &
     $-4.47 \pm 0.92$ &
     $-1.97 \pm 0.21$  \\
& Latent &
     $\boldsymbol{-1.34 \pm 0.33}$ &
     $-1.95 \pm 0.35$ &
     $-10.30 \pm 0.20$ &
     $-6.39 \pm 1.27$ &
     $-6.61 \pm 0.50$ &
     $-10.47 \pm 0.22$  \\ \midrule
{RNN-c} & Pred & $\boldsymbol{5.77 \pm 1.40}$ &
     $1.05 \pm 0.06$ &
     $0.81 \pm 0.03$ &
     $2.81 \pm 0.36$ &
     $0.86 \pm 0.02$ &
     $1.39 \pm 0.04$ \\
& Latent &
     $-2.30 \pm 0.61$ &
     $\boldsymbol{-2.05 \pm 0.32}$ &
     $-10.22 \pm 0.29$ &
     $-10.75 \pm 0.15$ &
     $-10.22 \pm 0.29$ &
     $-11.22 \pm 0.04$  \\ \midrule
\end{tabular}
\end{table*}


So far, we outlined a model that has the same conditional independence structure of the prior and is therefore incapable of modeling dependencies arising from colliders. Fortunately, the local auxiliary variables that we introduced in order to increase the flexibility of the local flows can assume a second role as "input ports" that can induce non-local couplings. In fact, by coupling their auxiliary variables we can induce statistical dependencies between any subset of variables in the model. Importantly, this coupling can take a very simple form since additional complexity can be modeled by the highway flows. We will now introduce a local auxiliary model with a non-trivial conditional independence structure inspired by the forward-backward algorithm. The graphical structure of the posterior probabilistic program is a sub-graph of the moralization of the prior graph~\cite{bishop2006pattern}. Therefore, in order to be able to capture all the dependencies in the posterior, it suffices to couple the auxiliary variables by reversing the arrows of the original prior model. In fact, all the parents of every node get coupled by the reversed arrows. As an example, consider the following probabilistic model (Fig.~\ref{fig: colliders}A):
$
   \rho_1({x}_1 \mid {x}_2, {x}_3) \rho_2({x}_2) \rho_3({x}_3 \mid {x}_4)  \rho_4({x}_4) \,.
$
We couple the auxiliary variables by reversing the arrows, as formalized in the graphical model
$
    p_1(\epsilon_1) p_2({\epsilon}_2, \mid {\epsilon}_1) p_3({\epsilon}_3, \mid {\epsilon}_1) p_4(\epsilon_4 \mid \epsilon_3)~.
$
The graphical structure of the resulting augmented model is shown in Fig.~\ref{fig: colliders}B. The auxiliary variables can be coupled in a very simple manner since the diffeomorphisms can add arbitrary complexity to the final joint density. Here we use a simple linear normal model:
\begin{equation}
    {\epsilon}_k \mid \vt{\upsilon}_k = \sum_{j=1}^K a_j \odot {\upsilon}_{j} + a_0 \odot \xi_k \,~,
    \label{eq: backward coupling}
\end{equation}
where $a_j \geq 0$ and $\sum_n a_n = 1$. In this expression, $\xi_k$ is a standard Gaussian vector and $\vt{\upsilon}_k = (\upsilon_1,...,\upsilon_K)$ is an array containing all the parents auxiliary variables (the auxiliaries of the children of $x_k$ in the original graph).

\subsection{Inference amortization}\label{sec: amor}

\begin{figure}[ht]
    \centering
    \includegraphics[width=0.35\textwidth]{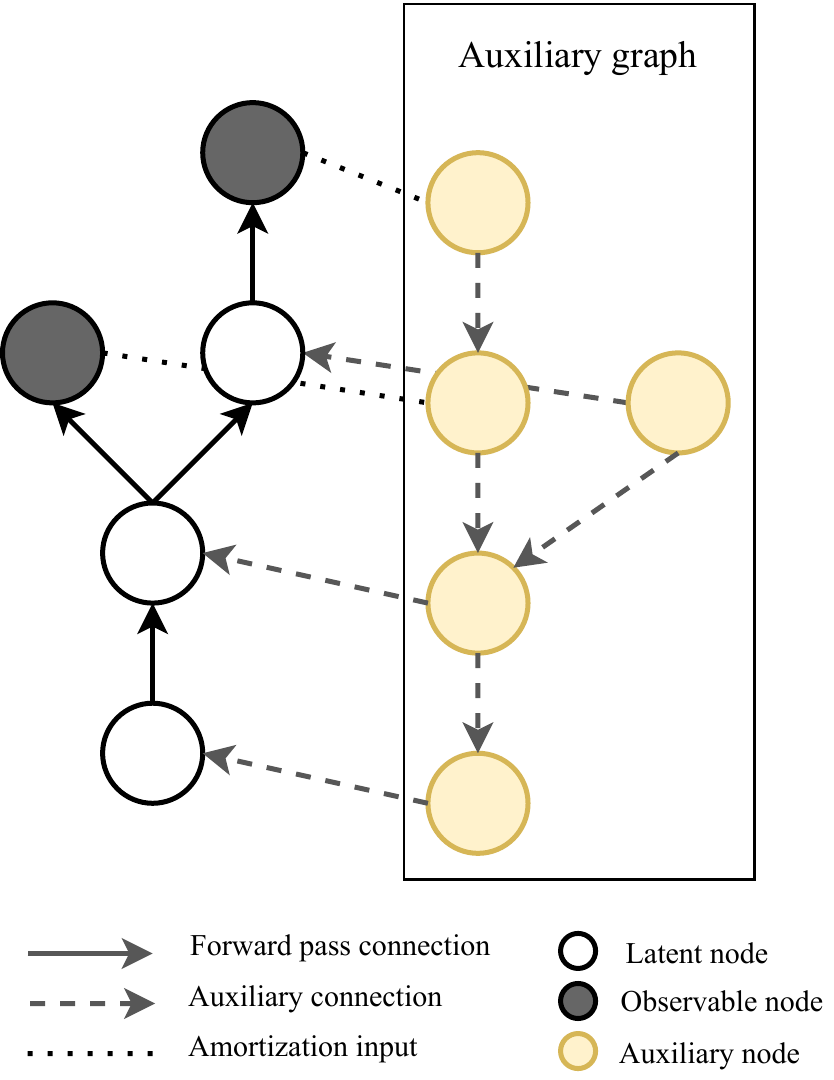}
   \caption{Visualization of the backward amortization algorithm.}
    \label{fig: amortization}
\end{figure}

We will now discuss a strategy to implement automatic inference amortization in cascading flows models. The problem of inference amortization is closely connected to the problem of modeling collider dependencies as the latter arise from the backward messages originating from observed nodes~\cite{bishop2006pattern}. It is therefore not a coincidence that we can implement both collider dependencies and amortization by shifting the statistics of the local auxiliary variables. This results in a very parsimonious amortization scheme where data streams are fed locally to the various nodes and propagate backward through the linear network of auxiliary Gaussian variables. We denote the set of auxiliary variables associated to the children of $x_k$ in the input probabilistic program as $\vt{\upsilon}_k$. Furthermore, we denote the observed value of $x_k$ as $y_k$ and the set of observed values of the $j$-th child of  $x_k$ as $\vt{y}^k = (y^k_1,...,y^k_K)$. The amortized auxiliary model can then be obtained as a modification of Eq.~\ref{eq: backward coupling}:
$
     p_0({\epsilon}_k \mid \vt{\upsilon}_k) = \mathcal{N}\!\left(m_k , \sigma^2 I \right)~
$
with
$
   {\epsilon}_k \mid \vt{\upsilon}_k = \mathcal{B}^{(k)}[y_k] + \sum_{j=1}^K a_j \odot {\upsilon}_{j} + a_0 \odot \xi_k \,~,
$
where $\mathcal{B}^{(k)}$ are learnable affine transformations (i.e. linear layers in deep learning lingo).

\section{Experiments}
Our experiments are divided into three sections. In the first section we focus on highly structured timeseries problems exhibiting linear and non-linear dynamics. Bayesian problems of this nature have relevance for several scientific fields such as physics, chemistry, biology, neuroscience and population dynamics. In the second section we compare the capacity of several variational programs to model dependencies arising from colliders in the graphical structure. In fact, the inability of modeling collider dependencies was the main limitation of the original convex-update ASVI family~\cite{ambrogioni2020automatic}. Finally, in the third section we test the performance of the automatic amortization scheme. All the code used in the experiments is included in the supplementary material and documented in Supplementary A. \textbf{Architectures}: In all experiments, the CF architectures were comprised of three highway flow blocks with softplus activation functions in each block except for the last which had linear activations. CF programs use an independent network of this form for each latent variable in the program. Each variable was supplemented with $10$ auxiliary variables, the width of each network was therefore equal to the dimensionality of the variable plus $10$. Weights and biases were initialized from centered normal distributions with scale $0.01$. The $\lambda$ variable was defined independently for each network as the logistic sigmoid of a leanable parameter $l$, which was initialized as $4$ in order to keep the variational program close to the input program. For each variable, the posterior auxiliary distributions $r(\epsilon)$ (see Eq.~\ref{eq: mixture ELBO}) were spherical normals parameterized by mean and standard deviation (i.e. Gaussian mean field).

\begin{figure}[ht]
    \centering
    \includegraphics[width=0.45\textwidth]{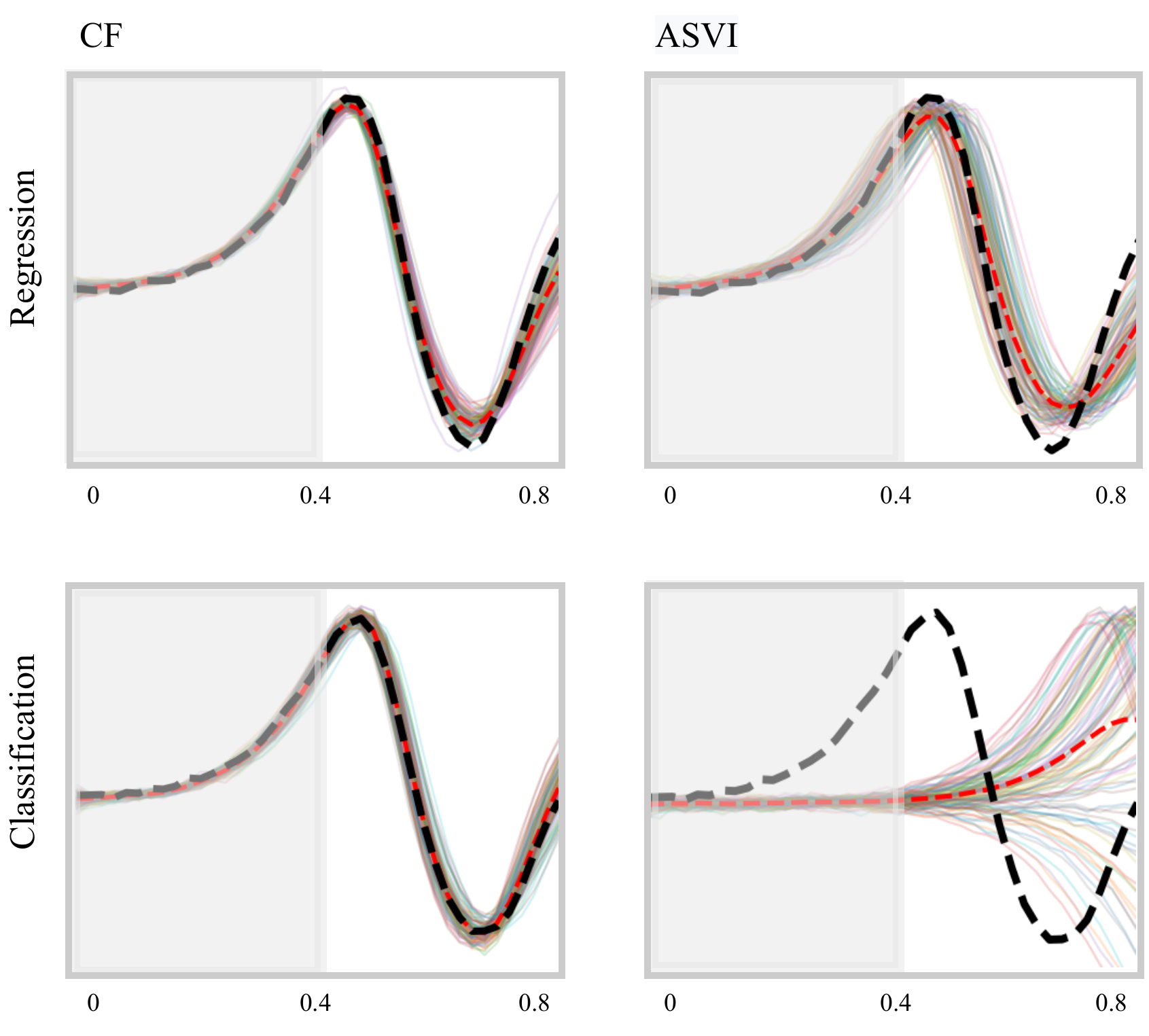}
   \caption{Example of qualitative results in the Lorentz system (first coordinate). The gray shaded area is observed through noisy measurements. The dashed lines are the ground-truth timeseries while the thin colored lines are samples from the trained variational program.}
    \label{fig: lorentz}
\end{figure}
\begin{table*}[]\label{tab: binary tree} \centering
\caption{Multivariate latent log-likelihood (forward KL) of variational binary tree models. Error are SEM estimated over 15 repetitions.}
\begin{tabular}{|l|l|l|l|l|l|}
\hline
         & CF        & ASVI & MF        & GF & MN \\ \hline
Linear-2 & $\boldsymbol{1.24 \pm 0.01}$  &  $0.49 \pm 0.05$    &   $0.6 \pm 0.01$        &  $0.97 \pm 0.01$  &  $1.00 \pm 0.01$  \\ \hline
Linear-4 & $5.42 \pm 0.35$ &  $3.56 \pm 0.26$    &  $1.68 \pm 0.02$         & $\boldsymbol{6.56 \pm 0.04}$   & $6.52 \pm 0.01 $   \\ \hline
Tanh-2  & $\boldsymbol{1.20 \pm 0.01}$  & $-7.86 \pm 3.47$ & $0.24 \pm 0.01 $    & $0.90 \pm 0.01$ &   $1.15 \pm 0.01$  \\ \hline
Tanh-4  & $3.29 \pm 3.17$ &  $-55.44 \pm 4.53$    &  $-43.24 \pm 8.87$         &  $\boldsymbol{10.32 \pm 0.05}$  & $5.37 \pm 2.61$   \\ \hline
\end{tabular}
\end{table*}

\subsection{Timeseries analysis}
 We consider discretized SDEs, defined by conditional densities of the form:
\begin{equation}
    \rho_t(x_{t+1}\mid x_t) = \mathcal{N}\left(x_{t+1}; \mu(x_{t}, t) \text{d}t , \sigma^2(x_{t}, t) \text{d}t \right)~,
    \label{eq: SDE}
\end{equation}
where $\mu(x_{t}, t)$ is the drift function and $\sigma(x_{t}, t)$ is the volatility function.
Specifically, we used Brownian motions (BR: $\mu(x) = x$, $\sigma^2(x) = 1$), Lorentz dynamical systems (LZ: $\mu(x_1,x_2,x_3) = (10(x_2 - x_1), x_1(28 - x_3) - x_2, x_1 x_2 - 8/3 x_3)$, $\sigma^2 = 2$), Lotka-Volterra population dynamics models (PD: $\mu(x_1, x_2) = (0.2 x_1 - 0.02 x_1 x_2, 0.1 x_1 x_2 - 0.1 x_2),\sigma^2 = 3$ ) and recurrent neural dynamics with randomized weights (RNN, see Supplementary B). The initial distributions were spherical Gaussians (see Supplementary B for the parameters). For each dynamical model we generated noisy observations using Gaussian emission models (denoted with an "r" in the tables): 
\begin{equation}
    p(y_t \mid x_t) = \mathcal{N}([x_t]_1, \sigma^2_{lk})
\end{equation}
or Bernoulli logistic models ("c" in the tables): 
\begin{equation}
    p(y_t \mid x_t) = \text{Bernoulli}(g(k [x_t]_1))~,
\end{equation}
where $[x_t]_1$ denotes the first entry of the vector $x_t$, $\sigma_{lk}$ is the standard deviation of the Gaussian emission model, $g(\cdot)$ is the logistic sigmoid function and $k$ is a gain factor. All the numeric values of the parameters in different experiments are given in Supplementary B. For each task we evaluate the performance of the cascading flows programs (CF) against a suite of baseline variational posteriors, including convex-update ASVI~\cite{ambrogioni2020automatic}, mean field (MF) and multivariate normal (MN)~\cite{kucukelbir2017automatic}. We also compare our new cascading flows family with the more conventional global normalizing flow (GF) approach where a single diffeomorphism transforms all the latent variables in the probabilistic program into a spherical normal distribution~\cite{rezende2015variational}. For the sake of comparison, we adopt an architecture identical to our highway flow model (including $10$ auxiliary variables) except for the absence of the highway gates. Note however that the global architecture has much greater width as it is applied to all variables in the program. Finally, we compare the performance of CF with a modified version of CF that does not use highway gates. Ground-truth multivariate timeseries $\tilde{\vt{x}} = (\tilde{x}_1,\dots,\tilde{x}_T)$ were sampled from the generative model together with simulated first-half observations $\vt{y}_{1:T/2} = (y_1,\dots,y_{T/2})$ and second-half observations $\vt{y}_{T/2:T} = (y_{T/2+1},\dots,y_{T})$. All the variational models were trained conditioned only on the first half observations. Performance was assessed using two metrics. The first metric is the average marginal log-probability of the ground-truth given the variational posterior:
$
    \frac{1}{T J} \sum_{t}^T \sum_j^J \log{q([\tilde{x}_t]_j \mid \vt{y}_{1:T/2})}~,
$
where $J$ is the dimensionality of the ground-truth vector $\tilde{x}_t$. The marginal posterior density $q([\tilde{x}_t]_j \mid \vt{y}_{1:T/2})$ was estimated from $5000$ samples drawn from the variational program using Gaussian kernel density estimation with bandwidth $0.9 \hat{\sigma} N^{-1/5}$~, where $\hat{\sigma}$ is the empirical standard deviation of the samples. Our second metric is $\log{p(\vt{y}_{T/2:T} \mid \vt{y}_{1:T/2})}$: the predictive log-probability of the ground-truth observations in the second half of the timeseries given the observations in the first half, estimated using kernel density estimation from $5000$ predicted observation samples drawn from the exact likelihood conditioned on the samples from the variational program. Each experiment was repeated $10$ times. In each repetition, all the variational programs were re-trained for $8000$ iterations (enough to ensure convergence in all methods) given new sampled ground-truth timeseries and observations. The reported results are the mean and SEM of the two metrics over these $10$ repetitions. \textbf{Results:} The results are given in Table 1. The cascading flows family (CF) outperforms all other approaches by a large margin in all non-linear tasks. Convex-update ASVI consistently has the second highest performance, easily outperforming the highly parameterized global normalizing flow (GF). This result is consistent with~\cite{ambrogioni2020automatic}, where ASVI was shown to outperform inverse autoregressive flow architectures~\cite{kingma2016improved}. Finally, the CF model without highway gates (CF non-res) has very low performance, proving the crucial importance of our new highway flow architecture. A qualitative comparison (LZ-r and LZ-c) between CF and ASVI is shown in Fig.~\ref{fig: lorentz}.

\subsection{Binary tree experiments}
In our second experiment we test the capability of the variational programs to model statistical dependencies arising from colliders. We tested this in a Gaussian binary tree model where the mean of a scalar variable $x_j^d$ in the $d$-th layer is a function of two variables in the $d-1$-th layer: 
$
    x_j^d \sim \mathcal{N}\!\left(\text{link}(\pi_{j1}^{d-1}, \pi_{j2}^{d-1}), \sigma^2 \right),
$
where $\pi_{j1}^{d-1}$ and $\pi_{j2}^{d-1}$ are the two parents of $x_j^d$. All the $0$-th layer variables have zero mean. The inference problem consists of observing the only variable in the last layer and computing the posterior of all the other variables. We considered a setting with linear coupling $\text{link}(x,y) = x - y$ and tanh coupling $\text{link}(x,y) = \tanh{(x)} - \tanh{(y)}$ with $2$ and $4$ layers. Performance was estimated using the posterior log-probability of the ground-truth sample estimated using a multivariate Gaussian approximation that matches the empirical mean and covariance matrix of $5000$ samples drawn from the trained variational program. We compared performance with ASVI, mean-field (MF), global flow (GF) and multivariate normal (MN). Note that neither ASVI nor MF can model collider dependencies. Full details are given in Supplementary C. The reported results are the mean and SEM of the two metrics over $15$ repetitions. \textbf{Results:} The results are given in Table 2. The CF model always outperforms ASVI and MF, proving that it can capture collider dependencies. CF has the highest performances in shallow models but lower performance than GF and MN in deeper models. 
\subsection{Inference amortization}
Finally, we evaluate the performance of the amortization scheme in non-linear timeseries inference problems. We used the population dynamics (PD) and the Lorentz (LZ) systems. The CF program was amortized using the approach described in Sec.~\ref{sec: amor}. The MF and normalizing flow baselines were amortized using inference networks (see Supplementary D). All the details were identical to the timeseries experiments except for the fact that all time points were observed. Performance was again quantified as the average marginal log-probability of the ground-truth kernel density estimated from the samples from the amortized variational programs. Each amortized program was trained once and tested on $50$ new sampled timeseries/observations pairs. The reported results are the mean and SEM of these $50$ repetitions. \textbf{Results:} The results are reported in Table~3. Like in the non-amortized case, CF greatly outperforms both MF and GF in both timeseries inference problems.

\begin{table}[]\label{tab: amortized}
\caption{Latent log-likelihood (forward KL) of amortized timeseries models. Error are SEM estimated over 50 repetitions.}
\begin{tabular}{|l|l|l|l|}
\hline
         & CF & MF & GF \\ \hline
PD & $\boldsymbol{-2.8 \pm 0.2}$  &  $-8.0 \pm 0.2$    &   $-5.1 \pm 0.2$  \\ \hline
LZ & $\boldsymbol{-2.7 \pm 0.2}$  &  $-10.0 \pm 0.2 $    &   $-9.1 \pm 0.2$   \\ \hline
\end{tabular}
\end{table}

\section{Discussion}
Our CF approach solves the two main limitations of ASVI as it can model collider dependencies and can be amortized automatically. We show that cascading flows programs have extremely high performance in structured inference problems. However, their ability to capture deep collider dependencies is lower than normalizing flows. In our opinion, designing principled ways to model collider dependencies is a promising research direction. 


\clearpage
\bibliographystyle{unsrtnat}
\bibliography{ref.bib}

\appendix 

\onecolumn

\section{Experiments}
The experiments where implemented in Python 3.8 using the packages PyTorch, Numpy and Matplotlib. The implementation code is contained in the "modules" folder. The probabilistic program objects are defined in the file "models.py" and the deep architectures are in "networks.py". The experiments can be run from the files "timeseries experiment.py", "collider linear experiments.py", "collider tanh experiments.py" and "amortized experiments.py" in the "experiments" folder.

\section{Models}
In the experiments, we use the following dynamical models. 

\subsection{Brownian motion (BR)}

\begin{tabular}{ll}
     Dimensionality: &  $J = 1$\\
     Drift:  & $\mu(x) = x$ \\ 
     Diffusion:  & $\sigma^2(x) = 1$ \\ 
     Initial density:  & $p(x_0) = \mathcal{N}(0, 1)$ \\ 
     Time horizon:  & $T = 40$ \\ 
     Time step:  & $\text{d}t = 1$ \\ 
     Regression noise:  & $\sigma_{\text{lk}} = 1$ \\ 
     Classification gain:  & $k = 2$ 
\end{tabular}

\subsection{Lorentz system (LZ)}

\begin{tabular}{ll}
     Dimensionality: &  $J = 3$\\
     Drift:  & $\mu(x_1,x_2,x_3) = (10(x_2 - x_1), x_1(28 - x_3) - x_2, x_1 x_2 - 8/3 x_3)$ \\ 
     Diffusion:  & $\sigma^2(x) = 2^2$ \\ 
     Initial density:  & $p(x_0) = \mathcal{N}(3, 20^2)$ \\ 
     Time horizon:  & $T = 40$ \\ 
     Time step:  & $\text{d}t = 1$ \\ 
     Regression noise:  & $\sigma_{\text{lk}} = 3$ \\ 
     Classification gain:  & $k = 2$ 
\end{tabular}

\subsection{Population dynamics (PD)}

\begin{tabular}{ll}
     Dimensionality: &  $J = 2$\\
     Drift:  & $\mu(x_1,x_2,x_3) = (\text{ReLu}(0.2 x_1 - 0.02 x_1 x_2), \text{ReLu}(0.1 x_1 x_2 - 0.1 x_2))$ \\ 
     Diffusion:  & $\sigma^2(x) = 2$ \\ 
     Initial density:  & $p(x_0) = \mathcal{N}(0, 1)$ \\ 
     Time horizon:  & $T = 100$ \\ 
     Time step:  & $\text{d}t = 0.02$ \\ 
     Regression noise:  & $\sigma_{\text{lk}} = 3$ \\ 
     Classification gain:  & $k = 2$ 
\end{tabular}
\\
\\
Note: The ReLu activation was added to avoid instabilities arising from negative values.

\subsection{Recurrent neural network (RNN)}

\begin{tabular}{ll}
     Dimensionality: &  $J = 3$\\
     Drift:  & $\mu(\boldsymbol{x}) = \tanh(W_3 \tanh(W_2 \tanh(W_1 \boldsymbol{x} + \boldsymbol{b_1}) + \boldsymbol{b_2}))$ \\ 
     Diffusion:  & $\sigma^2(x) = 0.1^2$ \\ 
     Initial density:  & $p(x_0) = \mathcal{N}(0, 1)$ \\ 
     Time horizon:  & $T = 40$ \\ 
     Time step:  & $\text{d}t = 0.02$ \\ 
     Regression noise:  & $\sigma_{\text{lk}} = 1$ \\ 
     Classification gain:  & $k = 2$ 
\end{tabular}
\\
\\
Note: $W_1$ is a $2 \times 5$ matrix, $W_2$ is a $5 \times 5$ matrix, $W_3$ is a $5 \times 2$ matrix, $\boldsymbol{b_1}$ is a 2D vector and $\boldsymbol{b_2}$ is a 5D vector. All the entries of these quantities are sampled from normal distributions with mean $0$ and standard deviation (SD) $1$. These parameters we re-sampled for every repetition of the experiment.

\section{Collider dependencies}
We tested performance under collider dependencies using a Gaussian binary tree model where the mean of a scalar variable $x_j^d$ in the $d$-th layer is a function of two variables in the $d-1$-th layer: 
$$
    x_j^d \sim \mathcal{N}\!\left(\text{link}(\pi_{j1}^{d-1}, \pi_{j2}^{d-1}), \sigma^2 \right)
$$
where $\pi_{j1}^{d-1}$ and $\pi_{j2}^{d-1}$ are the two parents of $x_j^d$. In the linear experiment the link was $\text{link}(\pi_{j1}^{d-1}, \pi_{j2}^{d-1}) = \pi_{j1}^{d-1} - \pi_{j2}^{d-1}$ The SD $\sigma$ was $0.15$ and the initial SD was $0.2$. In non-linear experiment the link was $\text{link}(\pi_{j1}^{d-1}, \pi_{j2}^{d-1}) = \tanh{\pi_{j1}^{d-1}} - \tanh{\pi_{j2}^{d-1}}$. The SD $\sigma$ was $0.05$ and the initial SD was $0.1$.

\section{Amortized experiments}
The amortized experiments had the same details and parameters as the corresponding timeseries experiment. We use the following inference network for the MF and GF baselines (PyTorch code):
    \begin{center}
    \includegraphics[width=1.\textwidth]{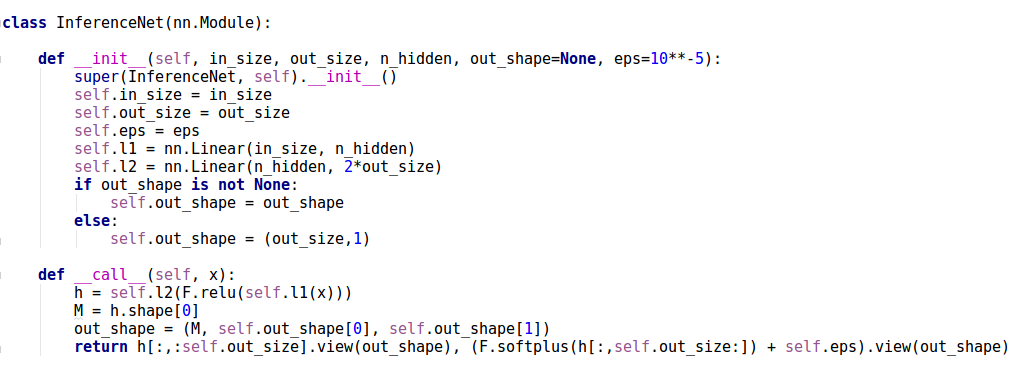}
    \end{center}
In the case of GF, the network provided means and scales of the pre-transformation normal variables.

\end{document}